\documentclass[preprint,12pt, a4paper]{elsarticle}

\usepackage{amssymb}
\usepackage{hyperref}

\journal{Software Impacts}

\begin{document}

\begin{frontmatter}



\title{SemSegLoss: A python package of loss functions for semantic segmentation}


\author{Shruti Jadon}

\address{Sunnyvale, CA 95134}

\begin{abstract}
Image Segmentation has been an active field of research as it has a wide range of applications, ranging from automated disease detection to self-driving cars. In recent years, various research papers proposed different loss functions used in case of biased data, sparse segmentation, and unbalanced dataset. In this paper, we introduce SemSegLoss, a python package consisting of some of the well-known loss functions widely used for image segmentation. It is developed with the intent to help researchers in the development of novel loss functions and perform an extensive set of experiments on model architectures for various applications. The ease-of-use and flexibility of the presented package have allowed reducing the development time and increased evaluation strategies of machine learning models for semantic segmentation. Furthermore, different applications that use image segmentation can use SemSegLoss because of the generality of its functions. This wide range of applications will lead to the development and growth of AI across all industries.
\end{abstract}


\end{frontmatter}

\noindent
\textbf{Introduction}\\

In recent years, Deep Learning has transformed multiple industries ranging from software to manufacturing. The medical community has also benefited from Deep Learning. There have been multiple innovations in disease classification, for example, lesion segmentation\cite{jadonspie20} using U-Net and cancer detection using SegNet. Image segmentation is one of the crucial contributions of the deep learning community. Image Segmentation can be defined as a classification task on the pixel level. An image consists of various pixels, and these grouped pixels define different elements in an image. A method of classifying these pixels into elements is called semantic image segmentation. The choice of loss/objective function is critical while designing complex image segmentation-based deep learning architectures as they instigate the learning process of the algorithm. Therefore, since 2012, researchers have experimented with various domain-specific loss functions to improve the model's performance on their datasets. This paper has introduced SemSegLoss, a python-based package consisting of some well-known loss functions widely used for image segmentation. Our implementation is available at GitHub: https://GitHub.com/shruti-jadon/Semantic-Segmentation-Loss-Functions.

\begin{table}[htbp]
\caption{Types of Semantic Segmentation Loss Functions \cite{jadon2020survey}}
\begin{center}
\begin{tabular}{|c|c|}
\hline
\textbf{Type} & {\textbf{Loss Function}} \\
\hline
Distribution-based Loss & Binary Cross-Entropy\cite{yi2004automated} \\
& Weighted Cross-Entropy\cite{pihur2007weighted}\\
& Balanced Cross-Entropy\cite{xie2015holistically}\\
& Focal Loss\cite{lin2002focal}\\
& Distance map derived loss penalty term\cite{caliva2019distance} \\
\hline
Region-based Loss & Dice Loss\cite{sudre2017generalised}\\
& Sensitivity-Specificity Loss{hashemi2018asymmetric}\\
& Tversky Loss\cite{salehi2017tversky}\\
& Focal Tversky Loss\cite{abraham2019novel} \\
& Log-Cosh Dice Loss\cite{jadon2020survey}\\
\hline
Boundary-based Loss & Hausdorff Distance loss\cite{karimi2019reducing}\\
& Shape aware loss\cite{hayder2016shape}\\
\hline
Compounded Loss & Combo Loss\cite{taghanaki2019combo} \\
& Exponential Logarithmic Loss\cite{wong20183d}\\
\hline
\end{tabular}
\label{tab2}
\end{center}
\end{table}

\section{Impact Overview}
SemSegLoss have a wide variety of application ranging from medicine to self-driving. While creating an optimal deep learning model, it is crucial to select the correct objective function (also known as loss functions). Therefore, it is crucial to have such frameworks that facilitate the research and development of these objective functions. However, to our knowledge, no other package provides a collection of loss functions for semantic segmentation. For this reason, all current researchers in this area are forced to spend hours searching for implementations or write their code. This process slows down the research of labs and sometimes hinders the experiment range. The main objective of SemSegLoss is to provide an easy way to experiment with various objective functions and determine the best possible approach, i.e., selecting loss function, which is giving the best performance and optimizing in fewer steps.

Furthermore, current segmentation-based loss functions are developed mainly by researchers with extensive knowledge of deep learning, linear algebra, and optimization. Nevertheless, thanks to the ease of use of SemSegLoss, more users will start research in this area. One of the strengths of this framework is that it is entirely written in Python, a programming language with an easy-to-understand syntax. In recent years a majority of machine learning-related development happened in Python. Moreover, it is also easy to modify or extend with not already implemented functionalities. Additionally, Python's flexibility facilitates the integration with modules written in other programming languages. SemSegLoss is a recently developed python package. It has received recognition from various researchers on GitHub. Simultaneously, SemSegLoss has been used to:
\begin{enumerate}
    \item \textbf{Create Novel Loss Functions:} SemSegLoss GitHub repo has been used to set-up the experiments for the claims of novel proposed loss functions such as Tilted Cross Entropy\cite{Szab2021TiltedCE} loss function, Mixed focal loss function\cite{yeung2021mixed}, and Soft Segmentation Loss Function\cite{gros2021softseg}
    \item \textbf{Perform Segmentation based experiments:} SemSegLoss code implementation is easy to follow which allowed applications to use the code implementation of loss functions for their segmentation based experiments. For example, cardiac function assessment in embryonic zebrafish\cite{naderi2021deep}, analyzing natural disaster aftermath from satellite images \cite{oludare2021semi}, and wildfire detection\cite{9377867}
\end{enumerate}
In all listed use-cases, the results obtained using SemSegLoss GitHub code implementation have provided researchers the ability to do choose the correct loss function to improve the models' performance.

\section{Functionalities and key features}
SemSegLoss currently has twelve widely used loss functions classified into four types as shown in table \ref{tab2}. If end-users want to analyze the efficiency of segmentation model architecture, they can use the diverse implementation of baseline Binary Cross-Entropy loss to Focal Tversky loss. Apart from the loss functions, the framework can also evaluate the model performance using different statistical analyses such as precision, recall, specificity, and Dice Coefficient. Using the SemSegLoss package, any programmer can experiment with loss functions and analyze the results using performance-based functionalities.

\section{Conclusions and further development}
SemSegLoss package provides easy-to-use loss functions to analyze and experiment with different approaches when developing an efficient segmentation-based model. Researchers and developers can take advantage of this package due to its simplicity. We intend to add more advanced loss function techniques such as Correlation Maximized, Structural Similarity Loss, and Distance map derived loss penalty term for medical-based image segmentation in future versions. We are also planning to upload the SemSegLoss package to PyPI to facilitate its distribution and use.

\section{Declaration of Competing Interest}
The authors declare that they have no known competing financial interests or personal relationships that could have appeared to influence the work reported in this paper.








\nocite{*}
\bibliography{main}{}

\begin{thebibliography}{10}

\bibitem{abraham2019novel}
Nabila Abraham and Naimul~Mefraz Khan.
\newblock A novel focal tversky loss function with improved attention u-net for
  lesion segmentation.
\newblock In {\em 2019 IEEE 16th International Symposium on Biomedical Imaging
  (ISBI 2019)}, pages 683--687. IEEE, 2019.

\bibitem{berman2017lovszsoftmax}
Maxim Berman, Amal~Rannen Triki, and Matthew~B. Blaschko.
\newblock The lovász-softmax loss: A tractable surrogate for the optimization
  of the intersection-over-union measure in neural networks, 2017.

\bibitem{caliva2019distance}
Francesco Caliva, Claudia Iriondo, Alejandro~Morales Martinez, Sharmila
  Majumdar, and Valentina Pedoia.
\newblock Distance map loss penalty term for semantic segmentation.
\newblock {\em arXiv preprint arXiv:1908.03679}, 2019.

\bibitem{gros2021softseg}
Charley Gros, Andreanne Lemay, and Julien Cohen-Adad.
\newblock Softseg: Advantages of soft versus binary training for image
  segmentation.
\newblock {\em Medical Image Analysis}, 71:102038, 2021.

\bibitem{hashemi2018asymmetric}
Seyed~Raein Hashemi, Seyed Sadegh~Mohseni Salehi, Deniz Erdogmus, Sanjay~P
  Prabhu, Simon~K Warfield, and Ali Gholipour.
\newblock Asymmetric loss functions and deep densely-connected networks for
  highly-imbalanced medical image segmentation: Application to multiple
  sclerosis lesion detection.
\newblock {\em IEEE Access}, 7:1721--1735, 2018.

\bibitem{hayder2016shape}
Zeeshan Hayder, Xuming He, and Mathieu Salzmann.
\newblock Shape-aware instance segmentation.
\newblock {\em arXiv preprint arXiv:1612.03129}, 2(5):7, 2016.

\bibitem{ho2019real}
Yaoshiang Ho and Samuel Wookey.
\newblock The real-world-weight cross-entropy loss function: Modeling the costs
  of mislabeling.
\newblock {\em IEEE Access}, 8:4806--4813, 2019.

\bibitem{jadon2020survey}
Shruti Jadon.
\newblock A survey of loss functions for semantic segmentation.
\newblock In {\em 2020 IEEE Conference on Computational Intelligence in
  Bioinformatics and Computational Biology (CIBCB)}, pages 1--7. IEEE, 2020.

\bibitem{jadonspie20}
Shruti Jadon, Owen~P. Leary, Ian Pan, Tyler~J. Harder, David~W. Wright, Lisa~H.
  Merck, and Derek~L. Merck.
\newblock {A comparative study of 2D image segmentation algorithms for
  traumatic brain lesions using CT data from the ProTECTIII multicenter
  clinical trial}.
\newblock In Po-Hao Chen and Thomas~M. Deserno, editors, {\em Medical Imaging
  2020: Imaging Informatics for Healthcare, Research, and Applications}, volume
  11318, pages 195 -- 203. International Society for Optics and Photonics,
  SPIE, 2020.

\bibitem{jiwani2021semantic}
Aatif Jiwani, Shubhrakanti Ganguly, Chao Ding, Nan Zhou, and David~M Chan.
\newblock A semantic segmentation network for urban-scale building footprint
  extraction using rgb satellite imagery.
\newblock {\em arXiv preprint arXiv:2104.01263}, 2021.

\bibitem{karimi2019reducing}
Davood Karimi and Septimiu~E Salcudean.
\newblock Reducing the hausdorff distance in medical image segmentation with
  convolutional neural networks.
\newblock {\em IEEE Transactions on medical imaging}, 39(2):499--513, 2019.

\bibitem{lin2002focal}
TY~Lin, P~Goyal, R~Girshick, K~He, and P~Doll{\'a}r.
\newblock Focal loss for dense object detection. arxiv 2017.
\newblock {\em arXiv preprint arXiv:1708.02002}, 2002.

\bibitem{moltz2020learning}
Jan~Hendrik Moltz, Annika H{\"a}nsch, Bianca Lassen-Schmidt, Benjamin Haas,
  A~Genghi, J~Schreier, Tomasz Morgas, and Jan Klein.
\newblock Learning a loss function for segmentation: A feasibility study.
\newblock In {\em 2020 IEEE 17th International Symposium on Biomedical Imaging
  (ISBI)}, pages 357--360. IEEE, 2020.

\bibitem{9377867}
Simone Monaco, Andrea Pasini, Daniele Apiletti, Luca Colomba, Paolo Garza, and
  Elena Baralis.
\newblock Improving wildfire severity classification of deep learning u-nets
  from satellite images.
\newblock In {\em 2020 IEEE International Conference on Big Data (Big Data)},
  pages 5786--5788, 2020.

\bibitem{naderi2021deep}
Amir~Mohammad Naderi, Haisong Bu, Jingcheng Su, Mao-Hsiang Huang, Khuong Vo,
  Ramses Seferino~Trigo Torres, J-C Chiao, Juhyun Lee, Michael~PH Lau, Xiaolei
  Xu, et~al.
\newblock Deep learning-based framework for cardiac function assessment in
  embryonic zebrafish from heart beating videos.
\newblock {\em arXiv preprint arXiv:2102.12173}, 2021.

\bibitem{oludare2021semi}
Victor Oludare, Landry Kezebou, Karen Panetta, and Sos Agaian.
\newblock Semi-supervised learning for improved post-disaster damage assessment
  from satellite imagery.
\newblock In {\em Multimodal Image Exploitation and Learning 2021}, volume
  11734, page 117340O. International Society for Optics and Photonics, 2021.

\bibitem{pan2019diagnostic}
Shiwen Pan, Wei Zhang, Wanjun Zhang, Liang Xu, Guohua Fan, Jianping Gong,
  Bo~Zhang, and Haibo Gu.
\newblock Diagnostic model of coronary microvascular disease combined with full
  convolution deep network with balanced cross-entropy cost function.
\newblock {\em IEEE Access}, 7:177997--178006, 2019.

\bibitem{pihur2007weighted}
Vasyl Pihur, Susmita Datta, and Somnath Datta.
\newblock Weighted rank aggregation of cluster validation measures: a monte
  carlo cross-entropy approach.
\newblock {\em Bioinformatics}, 23(13):1607--1615, 2007.

\bibitem{Puccio2016ThePC}
Benjamin Puccio, James~P. Pooley, John Pellman, Elise~C Taverna, and R.~Cameron
  Craddock.
\newblock The preprocessed connectomes project repository of manually corrected
  skull-stripped t1-weighted anatomical mri data.
\newblock {\em GigaScience}, 5, 2016.

\bibitem{Ribera2018WeightedHD}
Javier Ribera, David G{\"u}era, Yuhao Chen, and Edward~J. Delp.
\newblock Weighted hausdorff distance: A loss function for object localization.
\newblock {\em ArXiv}, abs/1806.07564, 2018.

\bibitem{salehi2017tversky}
Seyed Sadegh~Mohseni Salehi, Deniz Erdogmus, and Ali Gholipour.
\newblock Tversky loss function for image segmentation using 3d fully
  convolutional deep networks.
\newblock In {\em International Workshop on Machine Learning in Medical
  Imaging}, pages 379--387. Springer, 2017.

\bibitem{sudre2017generalised}
Carole~H Sudre, Wenqi Li, Tom Vercauteren, Sebastien Ourselin, and M~Jorge
  Cardoso.
\newblock Generalised dice overlap as a deep learning loss function for highly
  unbalanced segmentations.
\newblock In {\em Deep learning in medical image analysis and multimodal
  learning for clinical decision support}, pages 240--248. Springer, 2017.

\bibitem{Szab2021TiltedCE}
A.~Szab{\'o}, H.~J. Rad, and Siva-Datta Mannava.
\newblock Tilted cross entropy (tce): Promoting fairness in semantic
  segmentation.
\newblock {\em ArXiv}, abs/2103.14051, 2021.

\bibitem{taghanaki2019combo}
Saeid~Asgari Taghanaki, Yefeng Zheng, S~Kevin Zhou, Bogdan Georgescu, Puneet
  Sharma, Daguang Xu, Dorin Comaniciu, and Ghassan Hamarneh.
\newblock Combo loss: Handling input and output imbalance in multi-organ
  segmentation.
\newblock {\em Computerized Medical Imaging and Graphics}, 75:24--33, 2019.

\bibitem{wong20183d}
Ken~CL Wong, Mehdi Moradi, Hui Tang, and Tanveer Syeda-Mahmood.
\newblock 3d segmentation with exponential logarithmic loss for highly
  unbalanced object sizes.
\newblock In {\em International Conference on Medical Image Computing and
  Computer-Assisted Intervention}, pages 612--619. Springer, 2018.

\bibitem{xie2015holistically}
Saining Xie and Zhuowen Tu.
\newblock Holistically-nested edge detection.
\newblock In {\em Proceedings of the IEEE international conference on computer
  vision}, pages 1395--1403, 2015.

\bibitem{yeung2021mixed}
Michael Yeung, Evis Sala, Carola-Bibiane Sch{\"o}nlieb, and Leonardo Rundo.
\newblock A mixed focal loss function for handling class imbalanced medical
  image segmentation.
\newblock {\em arXiv preprint arXiv:2102.04525}, 2021.

\bibitem{yi2004automated}
Ma~Yi-de, Liu Qing, and Qian Zhi-Bai.
\newblock Automated image segmentation using improved pcnn model based on
  cross-entropy.
\newblock In {\em Proceedings of 2004 International Symposium on Intelligent
  Multimedia, Video and Speech Processing, 2004.}, pages 743--746. IEEE, 2004.

\bibitem{zhao2019correlation}
Shuai Zhao, Boxi Wu, Wenqing Chu, Yao Hu, and Deng Cai.
\newblock Correlation maximized structural similarity loss for semantic
  segmentation.
\newblock {\em arXiv preprint arXiv:1910.08711}, 2019.

\end{thebibliography}
\bibliographystyle{plain}


\section*{Required Metadata}
\label{}

\section*{Current code version}
\label{}

\begin{table}[!h]
\begin{tabular}{|l|p{6.5cm}|p{6.5cm}|}
\hline
\textbf{Nr.} & \textbf{Code metadata} & \textbf{Details} \\
\hline
C1 & Current code version & v.1.1 \\
\hline
C2 & Permanent link to code/repository used for this code version & For example: $https://GitHub.com/shruti-jadon/Semantic-Segmentation-Loss-Functions$ \\
\hline
C3  & Permanent link to Reproducible Capsule & $https://codeocean.com/capsule/1815956/tree$ \\
\hline
C4 & Legal Code License   & MIT \\
\hline
C5 & Code versioning system used & git \\
\hline
C6 & Software code languages, tools, and services used & Python \\
\hline
C7 & Compilation requirements, operating environments \& dependencies & \\
\hline
C8 & If available Link to developer documentation/manual & \\
\hline
C9 & Support email for questions & sjadon@umass.edu\\
\hline
\end{tabular}
\caption{Code metadata}
\label{} 
\end{table}

\end{document}